%% file: main.tex
\documentclass[10pt,twocolumn,letterpaper]{article}

\usepackage[pagenumbers]{cvpr}      %
\usepackage{graphicx}
\usepackage{amsmath}
\usepackage{amssymb}
\usepackage{booktabs}
\usepackage{subcaption}
\usepackage{multirow}
\usepackage{color}
\usepackage{tabularx,ragged2e,caption}
\usepackage{wrapfig,lipsum}
\usepackage{bbm}

\usepackage[pagebackref,breaklinks,colorlinks]{hyperref}

\usepackage[capitalize]{cleveref}
\crefname{section}{Sec.}{Secs.}
\Crefname{section}{Section}{Sections}
\Crefname{table}{Table}{Tables}
\crefname{table}{Tab.}{Tabs.}

\begin{document}

\title{Knowledge Distillation for 6D Pose Estimation\\by Aligning Distributions of Local Predictions}

\author{Shuxuan Guo\textsuperscript{1}, Yinlin Hu\textsuperscript{1}, Jose M. Alvarez\textsuperscript{2}, Mathieu Salzmann\textsuperscript{1} \\
$^1$CVLab, EPFL, Lausanne 1015, Switzerland \\
$^2$NVIDIA, Santa Clara, CA 95051, USA
\\
{\tt \small \{shuxuan.guo, yinlin.hu, mathieu.salzmann\}@epfl.ch} \quad{\tt \small josea@nvidia.com}}

\maketitle

\input{tex/0_abstract}

\input{tex/1_introduction}

\input{tex/2_related_work}

\input{tex/3_method}

\input{tex/4_experiment}

\input{tex/5_conclusion}

{\small
\bibliographystyle{ieee_fullname}
\bibliography{string,cvpr_2023}
}

\appendix
\section{Appendix}
\setcounter{figure}{0}
\setcounter{table}{0}
\setcounter{equation}{0}

\renewcommand{\thetable}{A\arabic{table}}
\renewcommand{\thefigure}{A\arabic{figure}}
\renewcommand{\theequation}{A.\arabic{equation}}
\input{tex/app_supp}

\end{document}

%% file: tex/0_abstract.tex
\begin{abstract}
\label{sec:abstract}

Knowledge distillation facilitates the training of a compact student network by using a deep teacher one. While this has achieved great success in many tasks, it remains completely unstudied for image-based 6D object pose estimation. 
In this work, we introduce the first knowledge distillation method driven by the 6D pose estimation task. To this end, we observe that most modern 6D pose estimation frameworks output local predictions, such as sparse 2D keypoints or dense representations, and that the compact student network typically struggles to predict such local quantities precisely. Therefore, instead of imposing prediction-to-prediction supervision from the teacher to the student, we propose to distill the teacher's distribution of local predictions into the student network, facilitating its training.
Our experiments on several benchmarks show that our distillation method yields state-of-the-art results with different compact student models and for both keypoint-based and dense prediction-based architectures.

\end{abstract}

%% file: tex/1_introduction.tex
\section{Introduction}
\label{sec:introduction}

Estimating the 3D position and 3D orientation, a.k.a. 6D pose, of an object relative to the camera from a single 2D image has a longstanding history in computer vision, with many real-world applications, such as robotics, autonomous navigation, and virtual and augmented reality. Modern methods that tackle this task~\cite{kendall2015posenet, xiang2017posecnn, peng2019pvnet, hu2019segpose, li2019cdpn, hu2021wdrpose, Wang_2021_GDRN, di2021SOpose, su2022zebrapose} all rely on deep neural networks.
The vast majority of them draw their inspiration from the traditional approach, which consists of establishing correspondences between the object's 3D model and the input image and compute the 6D pose from these correspondences using a Perspective-n-Point (PnP) algorithm~\cite{lepetit2009epnp,ke2017efficient, BarathM19progressivex,terzakis2020consistently} or a learnable PnP network. 
Their main differences then lie in the way they extract correspondences. While some methods predict the 2D image locations of sparse 3D object keypoints, such as the 8 3D bounding box corners~\cite{hu2019segpose, hu2020singlestagepose, hu2021wdrpose} or points on the object surface~\cite{peng2019pvnet}, others produce dense representations, such as 3D locations~\cite{Wang_2021_GDRN,di2021SOpose} or binary codes~\cite{su2022zebrapose}, from which the pose can be obtained. 

\begin{figure}[!tbp]
  \centering
    \includegraphics[width=0.7\linewidth]{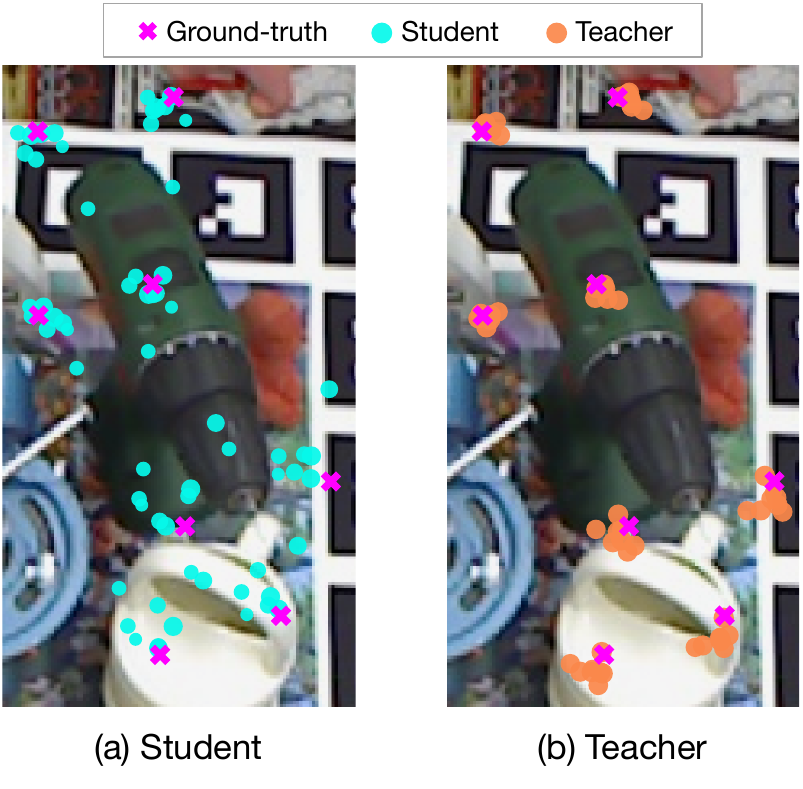}
    \vspace{-0.2cm}
    \caption{{\bf Student vs teacher keypoint predictions.} The large backbone of the teacher allows it to produce accurate keypoints, indicated by tight clusters. By contrast, because of its more compact backbone, the student struggles to predict accurate keypoints when trained with keypoint-to-keypoint supervision. We therefore propose to align the student's and teacher's keypoint \emph{distributions}. 
    }
    \label{fig:example}
\end{figure}

In any event, these methods rely on large models, which, while achieving impressive accuracy, are impractical deployment on embedded platforms and edge devices. As, to the best of our knowledge, no compact and efficient 6D pose estimation models have yet been proposed, a simple way to reduce the size of these networks consists of replacing their large backbones with much smaller ones. Unfortunately, this typically comes with a significant accuracy drop.  In this paper, we address this by introducing a knowledge distillation strategy for 6D pose estimation networks.

Knowledge distillation aims to transfer information from a deep teacher network to a compact student one. The research on this topic has tackled diverse tasks, such as image classification~\cite{hinton2015distilling,Zagoruyko2017AT, romero2014fitnets}, object detection~\cite{zhangimproveFKD, guo2021distilling, guo2021distillingDeFeat} and semantic segmentation~\cite{Liu_2019_CVPR, he2019knowledge}. While some techniques, such as feature distillation~\cite{romero2014fitnets,zhangimproveFKD, Zagoruyko2017AT,Heo_2019_ICCV}, can in principle generalize to other tasks, no prior work has studied knowledge distillation in the context of 6D pose estimation.

In this paper, we introduce a knowledge distillation method for 6D pose estimation motivated by the following observations. In essence, whether outputting sparse 2D locations or dense representations, the methods discussed above all produce multiple local predictions. We then argue that the main difference between the local predictions made by a deep teacher network and a compact student one consists in the accuracy of these individual predictions. Figure~\ref{fig:example} showcases this for sparse keypoint predictions, evidencing that predicting accurate keypoint locations with keypoint-to-keypoint supervision is much harder for the student than for the teacher. We therefore argue that knowledge distillation for 6D pose estimation should be performed not by matching the individual local predictions of the student and teacher but instead by encouraging the student and teacher \emph{distributions} of local predictions to become similar. This leaves more flexibility to the student and thus facilitates its training.

To achieve this, we follow an Optimal Transport (OT) formalism~\cite{villani2009optimal}, which lets us measure the distance between the two sets of local predictions. We express this as a loss function that can be minimized using a weight-based variant of Sinkhorn's algorithm~\cite{NIPS2013_sinkhorn}, which further allows us to exploit predicted object segmentation scores in the distillation process. Our strategy is invariant to the order and the number of local predictions, making it applicable to unbalanced teacher and student predictions that are not in one-to-one correspondence.

We validate the effectiveness of our approach by conducting extensive experiments on the popular LINEMOD~\cite{hinterstoisser2012model}, Occluded-LINEMOD~\cite{brachmann2014learning} 
and YCB-V~\cite{xiang2017posecnn} datasets with the SOTA keypoint-based approach WDRNet+.
Our prediction distribution alignment strategy consistently outperforms both a prediction-to-prediction distillation baseline and the state-of-the-art feature distillation method~\cite{zhangimproveFKD} using diverse lightweight backbones and architecture variations. Interestingly, our approach is orthogonal to feature distillation, and we show that combining it with the state-of-the-art approach of~\cite{zhangimproveFKD} further boosts the performance of student network.  
To show the generality of our approach beyond keypoint prediction, we then apply it to the SOTA dense prediction-based method, ZebraPose~\cite{su2022zebrapose}, to align the distributions of dense binary code probabilities. Our experiments evidence that this outperforms training a compact ZebraPose in a standard prediction-to-prediction knowledge distillation fashion.

Our main contributions can be summarized as follows. (i) We investigate for the first time knowledge distillation in the context of 6D pose estimation. %
(ii) We introduce an approach that aligns the teacher and student distributions of local predictions together with their predicted object segmentation scores. (iii) Our method generalizes to both sparse keypoints and dense predictions 6D pose estimation frameworks. (iv) Our approach can be used in conjunction with feature distillation to further boost the student's performance. 
We will release our code.

%% file: tex/2_related_work.tex
\section{Related Work}
\label{sec:related works}

\noindent \textbf{6D pose estimation.} With the great development and success of deep learning in computer vision~\cite{NIPS2012_imagenet, he2016deep, liu2016ssd, mrcnn, ErFNet2018, long2015fully}, many works have explored its use for 6D pose estimation. The first attempts~\cite{xiang2017posecnn, kendall2015posenet, kehl2017ssd} aimed to directly regress the 6D pose from the input RGB image. However, the representation gap between the 2D image and 3D rotation and translation made this task difficult, resulting in limited success. Therefore, most methods currently predict quantities that are closer to the input image space. 
In particular, several techniques jointly segment the object and predict either the 2D image locations of the corners of the 3D object bounding box~\cite{hu2019segpose,hu2020singlestagepose, hu2021wdrpose} or the 2D displacements from the cells' center of points on the object's surface~\cite{peng2019pvnet}. Instead of predicting such sparse 2D keypoints, other methods~\cite{li2019cdpn,Wang_2021_GDRN,di2021SOpose} output dense correspondences between the input image and the object 3D model, typically by predicting a 3D coordinate at every input location containing an object of interest. Recently, the state-of-the-art ZebraPose~\cite{su2022zebrapose} proposed to replace the prediction of 3D coordinates with that of binary codes encoding such coordinates, yet still producing dense predictions. In any event, the original backbones used by all the above-mentioned methods tend to be cumbersome, making them impractical for deployment in resource-constrained environments. However, replacing these backbones with more compact ones yields a significant performance drop. Here, we address this by introducing a knowledge distillation method for 6D pose estimation applicable to any method outputting local predictions, whether sparse or dense.

\begin{figure*}[tbp]
  \centering
    \includegraphics[width=0.9\textwidth]{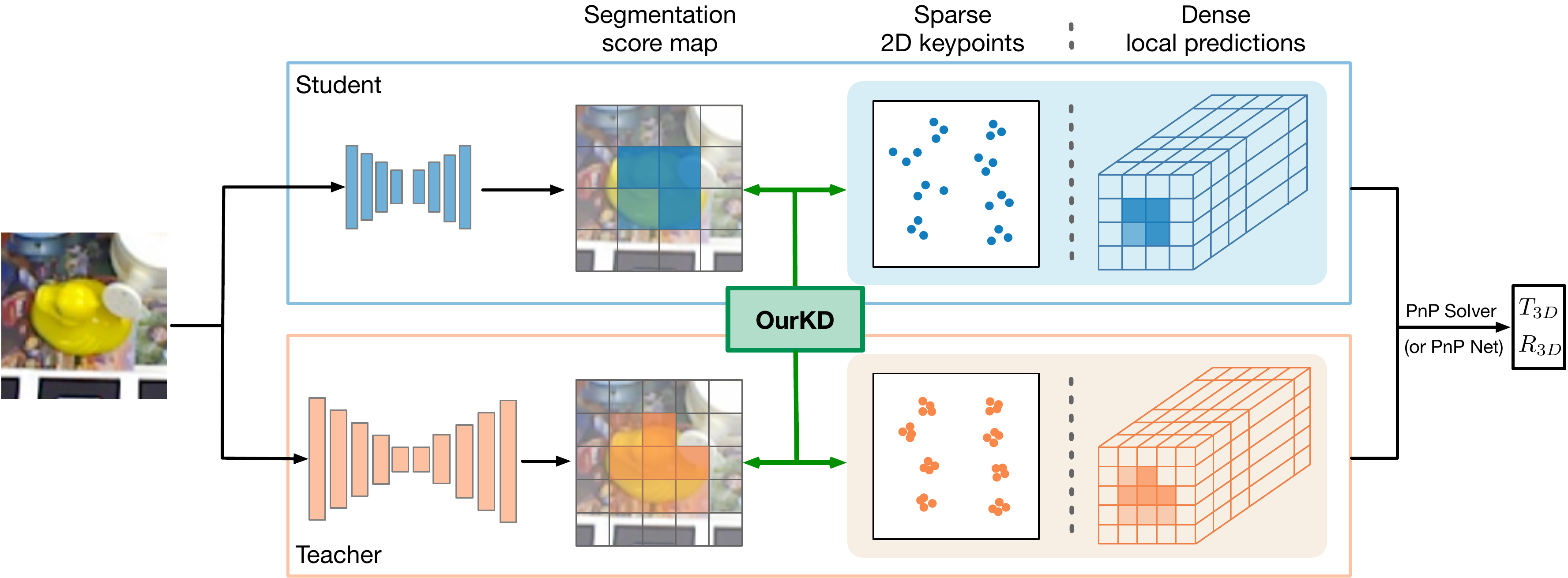}
    \vspace{-0.2cm}
    \caption{{\bf Overview of our method} (better viewed in color). 
    The teacher and student follow the same general architecture, predicting either sparse 2D keypoints or dense local predictions. Given an RGB input image, they output both a segmentation score map by classifying the individual cells in the feature map, and 2D keypoints voted by each cell as in~\cite{hu2021wdrpose}, or one prediction per cell, \eg, probabilities of 16D binary codes for ZebraPose~\cite{su2022zebrapose}. The local predictions, either sparse or dense, then form correspondences, which are passed to a PnP solver~\cite{lepetit2009epnp, BarathM19progressivex} or a PnP network~\cite{hu2020singlestagepose, Wang_2021_GDRN}
     to obtain the final 3D translation and 3D rotation. Instead of performing naive prediction-to-prediction distillation, we propose a strategy based on optimal transport that lets us jointly distill the teacher's local prediction distribution with the segmentation score map into the student.}
    \label{fig:framework}
\end{figure*}

\vspace{0.08cm}
\noindent \textbf{Knowledge distillation} has been proven effective to transfer information from a deep teacher to a shallow student in several tasks. This trend was initiated in the context of image classification, where Hinton~\etal~\cite{hinton2015distilling} guide the student's output using the teacher's class probability distributions, and Romero~\etal~\cite{romero2014fitnets}, Zagoruyko~\etal~\cite{Zagoruyko2017AT} and Tian~\etal~\cite{tian2019crd} encourage the student’s intermediate feature representations to mimic the teacher’s ones. Recently, many works have investigated knowledge distillation for other visual recognition tasks, evidencing the benefits of extracting task-driven knowledge. For example, in object detection, Zhang~\etal~\cite{zhangimproveFKD} adapt the feature distillation strategy of~\cite{romero2014fitnets} to object detectors; Wang~\etal~\cite{wang2019distillingFGFI} restrict the teacher-student feature imitation to regions around the positive anchors; Guo~\etal~\cite{guo2021distillingDeFeat} decouple the intermediate features and the classification predictions of the positive and negative regions; 
Guo~\etal\cite{guo2021distilling} distill detection-related knowledge from a classification teacher to a detection student. In semantic segmentation, Liu~\etal\cite{Liu_2019_CVPR} construct pairwise and holistic segmentation-structured knowledge to transfer. 
All of these works evidence that task-driven knowledge distillation boosts the performance of compact student models. 
Here, we do so for the first time for 6D object pose estimation.

\vspace{0.08cm}
\noindent \textbf{Optimal transport (OT)} has received a growing attention both from a theoretical perspective~\cite{villani2009optimal, NIPS2013_sinkhorn, santambrogio2015optimal} and for specific tasks, including shape matching~\cite{su2015optimal}, generative modeling~\cite{pmlr-v70-arjovsky17a}, domain adaptation~\cite{ot_for_da}, and model fusion~\cite{nips2020_ot_modelfusion}.
In particular, OT has the advantage of providing a theoretically sound way of comparing multivariate probability distributions without approximating them with parametric models. Furthermore, it can capture more useful information about the nature of the problem by considering the geometric or the distributional properties of the underlying space.
Our work constitutes the first attempt at using OT to align the student and teacher local prediction distributions for knowledge distillation in 6D pose estimation.

%% file: tex/3_method.tex
\section{Methodology}
\label{sec:method}
Let us now introduce our method to knowledge distillation for 6D pose estimation. As discussed above, 
we focus on approaches that produce local predictions, such as sparse 2D keypoints~\cite{hu2019segpose,peng2019pvnet,hu2020singlestagepose,hu2021wdrpose} or dense quantities~\cite{li2019cdpn, Wang_2021_GDRN, di2021SOpose, su2022zebrapose}.
In essence, the key to the success of such methods is the prediction of accurate local quantities. However, as shown in Figure~\ref{fig:example} for the keypoint case, the predictions of a shallow student network tend to be less precise than those of a deep teacher, i.e., less concentrated around the true keypoint locations in the figure, and thus yield less accurate 6D poses. 
Below, we first present a naive strategy to distill the teacher's local predictions into the student ones, and then introduce our approach.

\subsection{Naive Prediction-to-prediction Distillation}
\label{sec:naive}
The most straightforward way of performing knowledge distillation is to encourage the student's predictions to match those of the teacher.
In our context, one could therefore think of minimizing the distance between the local predictions of the teacher and those of the student. To formalize this, let us assume that the teacher and the student both output $N$ local predictions, i.e., that $N$ cells in the final 
feature maps participate in the prediction for the object of interest. Then, a naive distillation loss can be expressed as
\begin{equation}
\mathcal{L}_{naive-kd}(P^s, P^t)  = \sum_{i=1}^{N} \| P^s_{i} - P^t_{i} \|_{p} \;,
\end{equation}
where, $P^s_{i}$, resp. $P^t_{i}$, represent the student's, resp. teacher's, local predictions, and $p\in\{1,2\}$.

One drawback of this strategy comes from the fact that the teacher and student network may disagree on the number of local predictions they make. For example, as illustrated in Figure~\ref{fig:framework} for the keypoint case, the number of cells predicted to belong to the object by the student and the teacher may differ. This can be circumvented by only summing over the $\tilde{N}\leq N$ cells that are used by both the teacher and the student.
However, the distillation may then be suboptimal, as some student's predictions could potentially be unsupervised by the teacher.
Furthermore, and as argued above, a compact student tends to struggle when trained with prediction-to-prediction supervision, and such a naive KD formulation still follows this approach. Therefore, and as will be shown in our experiments, this naive strategy often does not outperform the direct student training, in particular in the sparse 2D keypoints scenario. Below, we therefore introduce a better-suited approach.

\subsection{Aligning the Distributions of Local Predictions}

In this section, we first discuss our general formulation, and then specialize it to sparse keypoint prediction and dense binary code prediction.
As discussed above and illustrated in Figure~\ref{fig:framework}, the number of student local predictions $N^s$ may differ from that of teacher local predictions $N^t$, preventing a direct match between the individual teacher and student predictions. To address this, and account for the observation that prediction-to-prediction supervision is ill-suited to train the student, we propose to align the \emph{distributions} of the teacher and student local predictions. We achieve this using optimal transport, which lets us handle the case where $N^s \neq N^t$. Formally, to allow the number of student and teacher predictions to differ, we leverage Kantorovich's relaxation~\cite{kantorovitch1958} of the transportation problem.

Specifically, assuming that all the local predictions have the same probability mass, i.e., $\frac{1}{N^t}$ for the teacher predictions and $\frac{1}{N^s}$ for the student ones, we derive a distillation loss based on Kantorovich's optimal transport problem as
\begin{equation}
    \vspace{-0.1cm}
    \begin{split}
     \bar{\mathcal{L}}_{kd} & (P^s, P^t; \pi)  =    \min _{\pi} \sum_{i=1}^{N^s}\sum_{j=1}^{N^t}\pi_{ij}\|P^s_{i} - P^t_{j}\|_p \; \\
    \text{s.t. \ \ } &\forall i, \;\; \sum_{j=1}^{N^t} \pi_{ij}=\frac{1}{N^s}\;, \ \ \forall j, \;\; \sum_{i=1}^{N^s} \pi_{ij} = \frac{1}{N^t}\;. 
    \end{split}
    \label{eq:ot_noscores}
    \vspace{-0.2cm}
\end{equation}
In our experiments, we found $p=2$ to be more effective than $p=1$ and thus use the $\ell_2$ norm below.

The above formulation treats all local predictions equally. However, different predictions coming from different cells in the feature maps might not have the same degree of confidence. In particular, this can be reflected by how confident the network is that a particular cell contains the object of interest, or, in other words, by a segmentation score predicted by the network. Let $\alpha^s_i$ denote such a score for cell $i$ in the student network, and $\alpha^t_j$ a similar score for cell $j$ in the teacher network.
We then re-write our distillation loss as
\begin{equation}
    \begin{split}
     \tilde{\mathcal{L}}_{kd} & (P^s, P^t; \alpha^s, \alpha^t; \pi)  =    \min _{\pi} \sum_{i=1}^{N^s}\sum_{j=1}^{N^t}\pi_{ij}\|P^s_{i} - P^t_{j}\|_2 \; \\ 
        &  \text{s.t. \ \ } \forall i, \;\; \sum_{j=1}^{N^t} \pi_{ij}=\alpha^s_i\;, \ \ \forall j, \;\; \sum_{i=1}^{N^s} \pi_{ij} = \alpha^t_j\;. 
    \end{split}
    \label{eq:ot_scores}
    \raisetag{24pt}
    \vspace{-0.1cm}
\end{equation}
In essence, because this loss involves both the local predictions and the cell-wise segmentation scores, it distills jointly the correspondence-related quantities and the segmentation results from the teacher to the student.

To solve this optimal transport problem, we rely on Sinkhorn's algorithm~\cite{NIPS2013_sinkhorn}, which introduces a soft versions of the constraints via Kullback-Leibler divergence regularizers. This then yields the final distillation loss
\begin{equation}
    \begin{split}
    \mathcal{L}_{kd}  (P^s, P^t; \alpha^s, & \alpha^t;  \pi)  =    \min _{\pi} \sum_{i=1}^{N^s}\sum_{j=1}^{N^t}\pi_{ij}\|P^s_{i} - P^t_{j}\|_2 \\
     & + \varepsilon^2 \mathrm{KL}(\pi, \alpha^s \otimes \alpha^t) + \rho^2 \mathrm{KL}(\pi \mathbf{1}, \alpha^s) \\
     &  +\rho^2 \mathrm{KL}\left(\pi^{\top} \mathbf{1}, \alpha^t\right) \;, 
    \end{split}
    % \vspace{-0.1cm}
    \label{eq:distillation_loss}
    \raisetag{13pt}
\end{equation}
where $\alpha^s$ and $\alpha^t$ concatenate the segmentation scores for the student and the teacher, respectively. 
This formulation was shown to be amenable to fast parallel optimization on GPU platforms, and thus well-suited for deep learning~\cite{NIPS2013_sinkhorn,feydy2019interpolating}.

\subsubsection{Keypoint Distribution Alignment}

Let us now explain how we specialize the formulation in Eq.~\ref{eq:distillation_loss} to the case of a network predicting sparse keypoints. In particular, we consider the case of predicting the 2D locations of the 8 object bounding box corners~\cite{hu2019segpose,peng2019pvnet,hu2020singlestagepose,hu2021wdrpose}. In this case, we consider separate costs for the 8 individual keypoints, to prevent a 2D location corresponding to one particular corner to be assigned to a different corner. 

Let $C^s_{k}$ and $C^t_{k}$ denote the predictions made by the student and the teacher, respectively, for the $k^{th}$ 2D keypoint location. Then, we express our keypoint distribution distillation loss as
\begin{equation}
    \begin{split}
    \mathcal{L}^{kp}_{kd}  (\{C^s_k\}, & \{C^t_k\}; \alpha^s, \alpha^t; \{\pi^k\}) \\ 
     & =   \sum_{k=1}^{8} \mathcal{L}_{kd} (C^s_{k}, C^t_{k}; \alpha^s, \alpha^t; \pi^{k}) .
    \end{split}
    \label{eq:distillation_loss_kp}
    \vspace{-0.2cm}
\end{equation}
In our experiments, we normalize the predicted 2D keypoints by the image size to the $[0,1]^2$ space, and set $\varepsilon$ to $0.001$ and $\rho$ to 0.5 to handle outliers.

\subsubsection{Dense Binary Code Distribution Alignment}

To illustrate the case of dense local predictions, we rely on the ZebraPose~\cite{su2022zebrapose} formalism, which predicts a 16-dimensional binary code probability vector at each cell of the final feature map. To further encode a notion of location in this dense representation, we concatenate the $x$- and $y$-coordinate in the feature map to the predicted vectors. Handling such dense representations, however, comes at a higher computational cost and memory footprint than with the previous sparse keypoints. To tackle this, we therefore average pool them over a small square regions.

Formally, let $B^s$ and $B^t$ represent the average-pooled local augmented binary code probabilities predicted by the student and teacher, respectively. Then, we write our dense prediction distribution distillation loss as
\begin{equation}
    \begin{split}
    \mathcal{L}^{bc}_{kd} & (B^s, B^t; \alpha^s, \alpha^t; \pi)  =  \mathcal{L}_{kd} (B^s, B^t; \alpha^s, \alpha^t; \pi) .
    \end{split}
    \label{eq:distillation_loss_dense}
    % \vspace{-0.1cm}
\end{equation}
where $\alpha^s$ and $\alpha^t$ also represent the average-pooled segmentation scores for the student and teacher, respectively. In our experiments, we use a pooling size of $8 \times 8$. Furthermore, we set $\varepsilon$ to $0.0001$ and $\rho$ to 0.1 to handle the outliers over the dense predictions of the binary code probabilities.

\subsection{Network Architectures} 
\label{sec:method_arch}
Our approach can be applied to any network that output local predictions. In our experiments, we use WDRNet~\cite{hu2021wdrpose} for the sparse keypoint case and ZebraPose~\cite{su2022zebrapose} for the dense prediction one. 
WDRNet employs a feature pyramid to predict the 2D keypoint locations at multiple stages of its decoder network. These multi-stage predictions are then fused by an ensemble-aware sampling strategy, ultimately still resulting in 8 clusters of 2D locations, i.e., one cluster per 3D bounding box corner. To make the WDRNet baseline consistent with the state-of-the-art methods~\cite{li2019cdpn,Wang_2021_GDRN, di2021SOpose, su2022zebrapose}, we incorporate a detection pre-processing step that provides an image patch as input to WDRNet. We refer to this as WDRNet+. We will nonetheless show in our experiments that the success of our distillation strategy does not depend on the use of this detector.
ZebraPose constitutes the state-of-the-art 6D pose estimation method. It predicts a binary code at each location in the feature map, and uses these codes to build dense 2D-3D correspondences for estimating 6D pose. 

In our experiments, the teacher and student networks follow the same general architecture, only differing in their backbones. Note that different backbones may also yield different number of stages in the WDRNet+ feature pyramid, but our distribution matching approach to knowledge distillation is robust to such differences. 
To train our WDRNet+ and ZebraPose networks, we rely on the standard losses proposed in~\cite{hu2021wdrpose, su2022zebrapose}. When performing distillation to a student network, we complement these loss terms with our distillation loss of either Eq.~\ref{eq:distillation_loss_kp}, for the keypoint case, or Eq.~\ref{eq:distillation_loss_dense} for the dense binary code one. To implement the losses, we rely on the GeomLoss library~\cite{feydy2019interpolating}.

%% file: tex/4_experiment.tex
\section{Experiments}
\label{sec:experiments}

\input{tables/kd_lm_2}

In this section, we first discuss our experimental settings, and then demonstrate the effectiveness and generalization ability of our approach on three widely-adopted datasets, LINEMOD~\cite{hinterstoisser2012model}, Occluded-LINEMOD~\cite{brachmann2014learning} and YCB-V~\cite{xiang2017posecnn}.
Finally, we analyze different aspects of our method and evaluate it on variations of our architecture.

\subsection{Experimental Settings}

\noindent \textbf{Datasets}. We conduct experiments on the standard LINEMOD~\cite{hinterstoisser2012model}, Occluded-LINEMOD~\cite{brachmann2014learning} and YCB-V~\cite{xiang2017posecnn} 6D pose estimation benchmarks. The LINEMOD dataset contains around 16000 RGB images depicting 13 objects, with a single object per image. Following~\cite{eric_uncertainty}, we split the data into a training set containing around 200 images per object and a test set  containing around 1000 images per object. The Occluded-LINEMOD dataset was introduced as a more challenging version of LINEMOD, where multiple objects heavily occlude each other in each RGB image. It contains 1214 testing images. For training, following standard practice, we use the real images from LINEMOD together with the synthetic ones provided with the dataset and generated using physically-based rendering~\cite{hlmo2020bop}. YCB-V~\cite{xiang2017posecnn} is a large dataset containing 21 strongly occluded objects observed in 92 video sequences, with a total of 133,827 frames. 

\vspace{0.08cm}
\noindent \textbf{Networks}. For WDRNet+, we use DarkNet53~\cite{redmon2018yolov3} as backbone for the teacher model, as in the original WDRNet~\cite{hu2021wdrpose}. For the compact students, we experiment with different lightweight backbones, including DarkNet-tiny~\cite{yolo} and a further reduced model, DarkNet-tiny-H, containing half of the channels of DarkNet-tiny in each layer. For ZebraPose, we use the pre-trained models of~\cite{su2022zebrapose} with a ResNet34~\cite{he2016deep} backbone as teacher networks and use DarkNet-tiny as backbone for the student networks.

\vspace{0.08cm}
\noindent \textbf{Baselines.} We compare our method to the direct training of the student without any distillation (Student), the naive knowledge distillation strategy introduced in Section~\ref{sec:naive} (Naive-KD), and the state-of-the-art feature distillation method (FKD)~\cite{zhangimproveFKD}, which, although only demonstrated for object detection, is applicable to 6D pose estimation. For these baselines, we report the results obtained with the best hyper-parameter values.  Specifically, 
for FKD, the best distillation loss weight on all three datasets was 0.01; for Naive-KD, the best weight was 0.1, and the best norm was $p=1$ for DarkNet-tiny and $p=2$ for DarkNet-tiny-H, respectively.
For our method, the distillation loss was set to 5 for LINEMOD and to 0.1 for both Occluded-LINEMOD and YCB-V. 
With ZebraPose, we conduct experiments on Occluded-LINEMOD only because of its much larger computational cost, taking many more iterations to converge than WDRNet+ (380K VS 200K). We use a distillation weight of 1.0 for Naive-KD and of 100.0 for our method. We provide the results of the hyper-parameter search in the supplementary material.

\vspace{0.08cm}
\noindent \textbf{Evaluation metric.} We report our results using the standard ADD-0.1d metric. It encodes the percentage of images for which the average 3D point-to-point distance between the object model in the ground-truth pose and in the predicted one is less than 10\% of the object diameter. For symmetric objects, the point-to-point distances are computed between the nearest points. Note that, on LINEMOD, we report the results obtained using the ground-truth 2D bounding boxes to remove the effects of the pretrained detectors. On Occluded-LINEMOD and YCB-V, we report the results obtained with the same detector as in~\cite{Wang_2021_GDRN, di2021SOpose, su2022zebrapose} to evidence the effectiveness of our knowledge distillation method.

\subsection{Experiments with WDRNet+}

Let us first consider the case of 2D keypoints with WDRNet+. In this scenario, we compare our keypoint distribution alignment method with the Naive-KD and the state-of-the-art feature distillation FKD with multiple student architectures on all three datasets.

\input{tables/kd_occ_main}

\vspace{0.08cm}
\noindent \textbf{Results on LINEMOD.} 
We report the results of our method and the baselines for all classes of the LINEMOD dataset in Table~\ref{tab:ab_lm_2}
for DarkNet-tiny and DarkNet-tiny-H. While Naive-KD slightly improves direct student training with the DarkNet-tiny backbone, it degrades the performance with DarkNet-tiny-H. This matches our analysis in Section~\ref{sec:method}; the fact that the student's and teacher's active cells differ make keypoint-to-keypoint distillation ill-suited.

Both FKD and our approach boost the student's results, with a slight advantage for our approach. In particular the accuracy improvement is larger, i.e., 2.3 points, for the smaller DarkNet-tiny-H backbone, for which the gap between the student and the teacher performance is also bigger. Note that the improvement of our approach over the student is consistent across the 13 objects.
Interestingly, the types of distillation performed by FKD and by our approach are orthogonal; FKD distills the intermediate features while we distill the predictions. As such, the two methods can be used together. As can be seen from the table, this further boosts the results, reaching a margin over the student of 1.7 points and 2.9 points with DarkNet-tiny and DarkNet-tiny-H, respectively, and thus constituting the state of the art on the LINEMOD dataset for such compact architectures.

\input{tables/kd_ycbv}
\vspace{0.08cm}
\noindent \textbf{Results on Occluded-LINEMOD.}
Let us now evaluate our method on the more challenging Occluded-LINEMOD. Here, we use only FKD~\cite{zhangimproveFKD} as baseline and drop Naive-KD due to its inferior performance shown before. 
The results are provided in Table~\ref{tab:lmocc_main}. Our keypoint-based knowledge distillation method yields results on par with the feature-based FKD on average. Note, however that FKD requires designing additional adaptive layers to match the misaligned feature maps, while our method does not incur additional parameters. More importantly, jointly using our method with FKD achieves the best results with 4.0 points improvements over the baseline student model. For some classes, such as \emph{can}, \emph{eggbox} and \emph{holepuncher}, the boost surpasses 5 points.

\vspace{0.08cm}
\noindent \textbf{Results on YCB-V.}
The results on the large YCB-V datasets are provided in Table~\ref{tab:ycbv}.
Our method outperforms the baseline and FKD by 2.6 and 1.3 on average. Moreover, the performance is boosted to 19.2 with Ours+FKD. These results further evidence the effectiveness of our method.

\subsection{Experiments with ZebraPose}
\label{sec:exp-zebrapose}

In Table~\ref{tab:lmocc_zebrapose}, we show the effectiveness of our method when applied to the SOTA dense prediction network ZebraPose~\cite{su2022zebrapose}. We compare our knowledge distillation strategy with the Naive-KD. In this dense prediction case, Naive-KD improves the baselines. Nevertheless, as evidenced by the results, our approaches outperforms both the student network and Naive-KD by 1.9 and 0.7 on average, respectively. This shows the generality of our KD method based on the alignment of local prediction distributions. 

\input{tables/kd_occ_zebrapose}

\input{tables/ab_ots}
\begin{figure*}[tbp]
  \centering
    \includegraphics[width=0.78\textwidth]{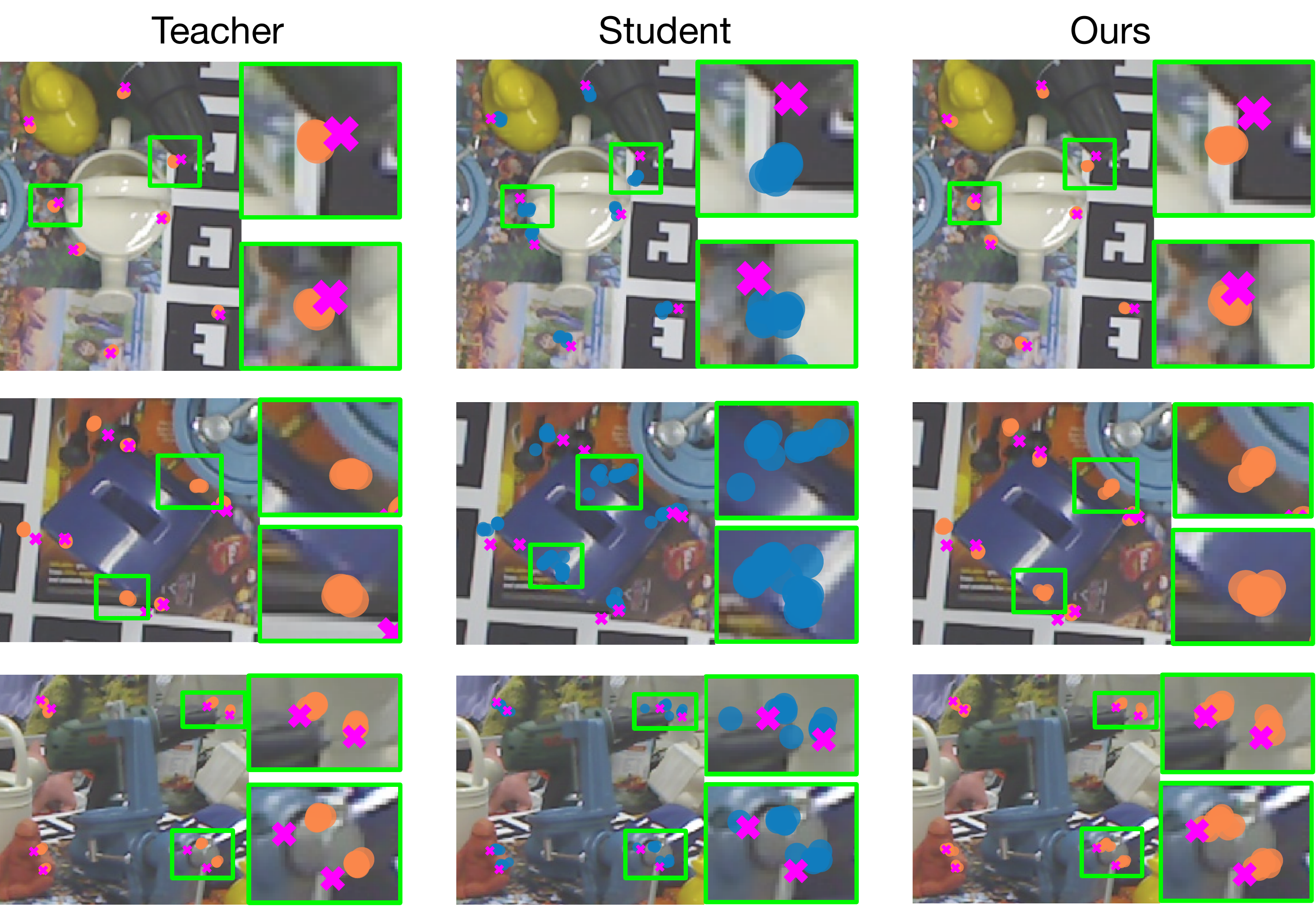}
    \vspace{-0.2cm}
    \caption{{\bf Qualitative Analysis} (better viewed in color). Comparison of the 2D keypoints predicted with our distilled model (3rd column with orange dots) and the baseline student model (2nd column with blue dots). With our distillation method, the model predicts tighter keypoint clusters, closer to the ground-truth corners (pink crosses) than the baseline model. Furthermore, our distilled model is able to mimic the teacher's keypoint distributions (1st column with orange dots). The light-green boxes highlight some keypoint clusters, which are also zoomed in on the side of the image.
    }
    \label{fig:vis_samples}
\end{figure*}

\subsection{Additional Analysis}
\label{sec:analysis}
Let us now further analyze the behavior of our knowledge distillation. 
The experiments in this section were performed using WDRNet+ on the LINEMOD dataset.

\vspace{0.08cm}
\noindent \textbf{With vs without segmentation scores.}
We compare the results of our approach without and with the use of the segmentation scores in the optimal transport formulation, i.e., Eq.~\ref{eq:ot_noscores}  vs Eq.~\ref{eq:ot_scores}. The comparison in Table~\ref{table:ab_ots} shows the benefits of jointly distilling the local predictions and the segmentation scores.

\vspace{0.08cm}
\noindent \textbf{Without detection pre-processing.}
Note that we incorporated the pre-processing detection step in WDRNet only because it has been shown to boost the pose estimation results.
However, the success of our knowledge distillation strategy does not depend on it. To demonstrate this, in the left portion of Table~\ref{tab:ab_lm_wdrs}, we report the results of our approach applied to the original WDRNet with a DarkNet-tiny backbone. As a matter of fact, the gap between direct student training and our approach is even larger (1.2 vs 2.1), showing the benefits of our approach on weaker networks.

\vspace{0.08cm}
\noindent \textbf{With a simple PnP network.}
In the right portion of Table~\ref{tab:ab_lm_wdrs}, we compare the results of our approach with those of the baselines on an architecture obtained by incorporating a simple PnP network at the end of WDRNet, following 
the strategy of
~\cite{hu2020singlestagepose}. With such an architecture, the 2D keypoint locations only represent an intermediate output of the network, with the PnP module directly predicting the final 3D translation and 3D rotation from them. As can be seen from these results, our distillation strategy still effectively boosts the performance of the student with this modified architecture, further showing the generality of our approach, which can distill keypoint-based knowledge both for PnP solvers and PnP networks.

\input{tables/ab_wdr_single_linemod}

\vspace{0.08cm}
\noindent \textbf{Qualitative analysis.}
We further provide visual comparisons of the predicted 2D keypoints distributions obtained with the baseline student model and with our distilled model on several examples from Occluded-LINEMOD. As shown in Figure~\ref{fig:vis_samples}, the predicted 2D keypoints clusters from our distilled models are closer to the ground-truth object corners than those of the baseline model. Furthermore, our distilled model mimics the teacher's keypoints distributions. 

\vspace{0.08cm}
\noindent \textbf{Limitations}. Because of the OT algorithm, training with our method comes with an overhead. 
Note, however, that we have observed this to have a negligible impact on the actual training clock time.
Furthermore, inference comes with no additional cost, and our distilled student model yields better performance. We have also observed that different classes benefit differently from distillation. This raises the possibility of designing class-wise distillation strategy, which we believe could be an interesting direction to explore in the future.

%% file: tables/kd_lm_2.tex
\begin{table*}[t]
    \centering
    \captionof{table}{\textbf{Results of DarkNet-tiny and DarkNet-tiny-H backbone on LINEMOD dataset with WDRNet+.} We report the ADD-0.1d for the baseline model, Naive-KD, FKD~\cite{zhangimproveFKD} and our KD method for each class. Our method not only outperforms Naive-KD and FKD, but can also be combined with FKD to obtain a further performance boost, yielding state-of-the-art results.}
    \label{tab:ab_lm_2}
    \vspace{-0.2cm}
    \resizebox{1.0\linewidth}{!}{
    \begin{tabular}{r |  c  | c | c  c  c c c | c | c  c   c  c c   } %
    \toprule
    \multirow{2}{*}{Class} & \multirow{2}{*}{Teacher} & \multicolumn{5}{c}{DarkNet-tiny}  &&\multicolumn{5}{c}{DarkNet-tiny-H}  \\
     \cmidrule{3-13}
    &    &   Student & Naive-KD & FKD & Ours  & Ours+$\dag$ & &  Student & Naive-KD & FKD & Ours  & Ours+$\dag$ \\ 
        \midrule
    Ape & 82.6  & 73.4 & 74.1 & 74.8 & 74.7 &  \textbf{76.2} && 65.4 & 64.1 & 68.4 & 69.4 &  \textbf{69.9}  \\ %
    Bvise       & 95.5  & 95.2 & 95.4 & 94.2 & 95.5 &  \textbf{96.7} & & 92.0 & 91.4 & 92.8 & \textbf{93.8} &   93.7    \\
    Cam         & 93.8  & 91.2 & 89.7 & 91.3 & 91.3 & \textbf{92.0} && 78.4 & 79.1 & 83.8 & 84.5 &   \textbf{84.5}                            \\
    Can         & 95.7  & 92.4 & 92.7 & \textbf{94.4} & 92.2 & 94.0 && 82.2 & 81.0 & 83.3 & 83.9 &   \textbf{83.9}     \\        
    Cat         & 92.0 & 87.2 & 85.0 & 87.5 & 88.4 & \textbf{88.6} && 81.5 & 78.7 & 80.7 & \textbf{81.8} &   81.6                             \\
    Driller     & 94.8  & 92.2 & 93.1 & 94.8 & 93.3 & \textbf{94.8} && 85.5 & 87.4 & \textbf{90.5} & 90.0 &   90.3                                \\
    Duck        & 76.0 & 70.9 & 74.4 & 73.6 & 73.5 & \textbf{74.7} & & 64.3 & 63.6 & 66.8 & 66.5 &  \textbf{68.9}  &  \\ %
    Eggbox$^*$  & 99.1 & 99.3 & 98.7 & 98.9 & 99.1 & \textbf{99.3} && 95.8 & 95.0 & 96.3 & 96.4 &  \textbf{96.4}                            \\
    Glue$^*$    & 96.4 & 97.2 & 97.1 & 96.2 & 97.7 & \textbf{97.7} && 90.7 & 91.2 & 91.0 & 91.9 &  \textbf{93.2} \\ %
    Holep       & 86.2  & 78.0 & 82.1 & 79.5 & \textbf{82.4} & 82.2 && 73.2 & 72.3 & \textbf{77.5} & 74.1 &  76.3                         \\
    Iron        & 93.6 & 92.1 & 92.1 & 91.4 & \textbf{93.5} & 93.2 && 86.3 & 86.3 & 87.6 & 88.7 &  \textbf{90.5}  \\ %
    Lamp        & 97.7  & 96.6 & 95.3 & 96.9 & \textbf{97.0} & 96.8 && 93.6 & 94.2 & 93.4 & \textbf{94.8} &  94.6 \\
    Phone       & 91.2 & 87.5 & 88.4 & 89.4 & 88.2 & \textbf{89.6} && 76.0 & 75.8 & \textbf{80.6} & 78.2 &  79.2                        \\
    \midrule
     \multirow{2}{*}{AVG. } & \multirow{2}{*}{91.9} & \multirow{2}{*}{88.7} & 89.1 & 89.4 & 89.9 & \textbf{90.4 } &&  \multirow{2}{*}{81.9} & 81.6 & 84.1 & 84.2 &  \textbf{84.8}  \\
     & & & ($\uparrow$ 0.4)& ($\uparrow$ 0.7) & ($\uparrow$ 1.2) & \textbf{($\uparrow$ 1.7)} && &($\downarrow$ 0.3) & $(\uparrow$ 2.2) & ($\uparrow$ 2.3) & \textbf{($\uparrow$ 2.9}) \\

    \bottomrule
    \multicolumn{13}{l}{\footnotesize{$\dag$ Ours+: Ours+FKD distills both the predictions and the intermediate feature maps.}}
    \end{tabular}
    }
\end{table*}

%% file: tables/kd_occ_main.tex
\begin{table}[!t]
    \centering
    \captionof{table}{\textbf{Results on OCC-LINEMOD with WDRNet+.} We report the ADD-0.1d for each class. Our method performs on par with FKD~\cite{zhangimproveFKD}, combining it with FKD yields a further performance boost.}
    \label{tab:lmocc_main}
    \vspace{-0.2cm}
    \resizebox{0.98\linewidth}{!}{
    \begin{tabular}{r |  c  |  c |  c  c c c  c } 
    \toprule
    Class   & Teacher    & Student  & FKD & Ours & Ours+  \\
    \midrule
    Ape          & 33.4      & 25.5      & 26.7      & 25.7      & \textbf{26.9}  \\
    Can          & 70.9      & 46.6      & 53.9      & 53.5      & \textbf{54.7}   \\
    Cat          & 45.1      & 31.4      & 31.1      & 32.2      & \textbf{32.9}   \\
    Driller       & 70.9      & 51.2      & 52.1      & 52.9      & \textbf{52.9}  \\
    Duck          & 27.0      & 22.5      & 25.3      & 25.7      & \textbf{27.0}  \\
    Eggbox$^*$   & 53.7      & 43.4      & 49.0      & 48.2      & \textbf{50.0}  \\
    Glue$^*$      & 70.7      & 54.5      & 55.6      & 55.8      & \textbf{56.9}  \\
    Holep         & 59.7      & 49.3      & 52.2      & 52.1      & \textbf{54.5}  \\
    \midrule
    \multirow{2}{*}{AVG. }         & \multirow{2}{*}{53.9}      & \multirow{2}{*}{40.5}      & 43.2      & 43.2  & \textbf{44.5} \\
    & & & ($\uparrow$ 2.7) & ($\uparrow$ 2.7) & \textbf{($\uparrow$ 4.0)} \\
    \bottomrule
    \end{tabular}
    }
\end{table}

%% file: tables/kd_ycbv.tex
\begin{table}[!t]
    \centering
    \captionof{table}{\textbf{Average results on YCB-V with WDRNet+}. Our method outperforms FKD~\cite{zhangimproveFKD} and further boosts the performance combining with it.}
    \label{tab:ycbv}
    \vspace{-0.2cm}
    \resizebox{0.82\linewidth}{!}{
    \begin{tabular}{c | c |  c    c  c } 
    \toprule
    Teacher  &  Student   & FKD  & Ours  & Ours+  \\
    \midrule
    \multirow{2}{*}{46.9}  & \multirow{2}{*}{16.1}       &  17.4      & \textbf{18.7}   & \textbf{19.2} \\
    & & ($\uparrow$ 1.3) & ($\uparrow$ 2.6) & \textbf{($\uparrow$ 3.1)} \\
    \bottomrule
    \end{tabular}
    }
\end{table}

%% file: tables/kd_occ_zebrapose.tex
\begin{table}[!t]
    \centering
    \captionof{table}{\textbf{Results on OCC-LINEMOD with ZebraPose~\cite{su2022zebrapose}} We report the ADD-0.1d for each class. Our method outperforms Naive-KD with ZebraPose, showing the generality of our approach to the dense prediction based method.}
    \label{tab:lmocc_zebrapose}
    \vspace{-0.2cm}
    \resizebox{0.92\linewidth}{!}{
    \begin{tabular}{r |  c | c |   c  c } 
    \toprule
    Class  & Teacher    & Student   & Naive-KD  & Ours   \\
    \midrule
    Ape     & 57.9      & 47.2      &  51.1     & \textbf{52.0} \\
    Can     & 95.0      & 93.2      &  93.5     & \textbf{94.2} \\
    Cat     & 60.6      & 53.1      &  53.9     & \textbf{55.2} \\
    Driller & 94.8      & 90.3      &  90.0     & \textbf{90.4}\\
    Duck    & 64.5      & 57.2      &  60.7     & \textbf{61.0} \\
    Eggbox$^*$ & 70.9   & 69.6      &  70.0     & \textbf{70.7} \\
    Glue$^*$ & 88.7     & 84.1      &  83.7     & \textbf{84.3} \\
    Holep  & 83.0       & 75.8      &  78.3     &\textbf{78.8}  \\                 

    \midrule
    \multirow{2}{*}{AVG. }    & \multirow{2}{*}{76.9}  &    \multirow{2}{*}{71.4}       &  72.6      & \textbf{73.3} \\
    & & & ($\uparrow$ 1.2) & \textbf{($\uparrow$ 1.9)} \\
    \bottomrule
    \end{tabular}
    }
\end{table}

%% file: tables/ab_ots.tex
\begin{table}[!t]
    \centering
    \captionof{table}{\textbf{Ablation study on LINEMOD: With vs without segmentation scores.}}
    \vspace{-0.2cm}
    \label{table:ab_ots}
    \resizebox{0.85\linewidth}{!}{
    \begin{tabular}{ r |  c  | c } 
	    \toprule
		Model               & \#Param(M)     & ADD-0.1d  \\ 
		\midrule
		WDRNet+(tiny)         &\multirow{3}{*}{8.5}     & 88.7  \\
		Ours-NoScores       &                         & 89.1  \\
		Ours                &                         & \textbf{89.9} \\
		\midrule
	   WDRNet+(tiny-H)       & \multirow{3}{*}{2.3}  & 81.9  \\
		Ours-NoScores       &                       & 83.1 \\
		Ours                &                       & \textbf{84.2}      \\
		\bottomrule
    \end{tabular} 
    }
\end{table}

%% file: tables/ab_wdr_single_linemod.tex
\begin{table}[!t]
    \centering
    \captionof{table}{\textbf{Evaluation under different network settings on LINEMOD.}
    We report the ADD-0.1d with the original WDRNet framework~\cite{hu2021wdrpose} and with an additional simple PnP network~\cite{hu2020singlestagepose}. Our method improves the performance of the student network in both settings.}
    \label{tab:ab_lm_wdrs}
    \vspace{-0.2cm}
    \resizebox{1.0\linewidth}{!}{
    \begin{tabular}{r |  c  | c  c   c|  c  | c c   } %
    \toprule
    \multirow{2}{*}{Class} &  \multicolumn{3}{c}{WDRNet}  &&\multicolumn{3}{c}{WDRNet + PnPNet}  \\
     \cmidrule{2-8}
                &   Teacher & Student & Ours & &  Teacher & Student & Ours \\ 
        \midrule
        Ape         & 70.3 & 41.2 & 43.0         & & 50.6 & 29.4 & 35.1 \\
        Bvis        & 94.2 & 81.5 & 86.1         & & 91.7 & 72.9 & 80.8 \\
        Cam         & 89.0 & 67.6 & 69.8         & & 90.5 & 56.1 & 73.3 \\
        Can         & 90.6 & 72.1 & 73.8         & & 88.3 & 57.5 & 75.9 \\
        Cat         & 87.1 & 54.3 & 61.5         & & 62.5 & 61.8 & 48.5 \\
        Driller     & 93.6 & 78.3 & 79.3         & & 87.1 & 68.6 & 71.9 \\
        Duck        & 64.5 & 35.9 & 39.6         & & 38.1 & 32.0 & 39.6 \\
        Eggbox$^*$  & 95.4 & 79.3 & 83.8         & & 99.3 & 91.8 & 96.6 \\
        Glue$^*$    & 93.4 & 83.4 & 82.7         & & 92.8 & 87.3 & 92.2 \\
        Holep       & 77.1 & 44.2 & 46.9         & & 70.9 & 46.4 & 49.9 \\
        Iron        & 90.9 & 75.8 & 75.1         & & 93.3 & 76.1 & 80.3 \\
        Lamp        & 96.3 & 84.8 & 86.8         & & 95.8 & 68.7 & 87.2 \\
        Phone       & 85.3 & 69.6 & 67.3         & & 92.3 & 57.0 & 76.6 \\
        \midrule
        \multirow{2}{*}{AVG. }    & \multirow{2}{*}{86.7 }  & \multirow{2}{*}{66.8 }  & \textbf{68.9}  &  & \multirow{2}{*}{81.0 } & \multirow{2}{*}{62.0 } & \textbf{69.8} \\
        & & & \textbf{($\uparrow$ 2.1)} & & & &  \textbf{($\uparrow$ 7.8)} \\
    \bottomrule
    \end{tabular}
    }
\end{table}

%% file: tex/5_conclusion.tex
\section{Conclusion}
\label{sec:conclusion}

We have introduced the first approach to knowledge distillation for 6D pose estimation. Our method is driven by matching the distributions of local predictions from a deep teacher network to a compact student one. We have formulated this as an optimal transport problem that lets us jointly distill the local predictions and the classification scores that segment the object in the image. Our approach is general and can be applied to any 6D pose estimation framework that outputs multiple local predictions. We have illustrated this with the sparse keypoint case and the dense binary code one. Our experiments have demonstrated the effectiveness of our method and its benefits over a naive prediction-to-prediction distillation strategy. Furthermore, our formalism is complementary to feature distillation strategies and can further boost its performance. In essence, our work confirms the importance of developing task-driven knowledge distillation methods, and we hope that it will motivate others to pursue research in this direction, may it be for 6D pose estimation or for other tasks.

%% file: tex/app_supp.tex
\subsection{Datasets and Codebase}

Here, we provide the details of the existing assets we used in our work, such as the LINEMOD~\cite{hinterstoisser2012model}, Occluded-LINEMOD~\cite{brachmann2014learning} and YCB-V~\cite{xiang2017posecnn} datasets, the GeomLoss library~\cite{feydy2019interpolating} and the original WDRNet~\cite{hu2021wdrpose} and ZebraPose~\cite{su2022zebrapose} codebase. All of them are open source and available for non-commercial academic research.

\textbf{LINEMOD~\cite{hinterstoisser2012model} and Occluded-LINEMOD~\cite{brachmann2014learning}}\footnote{~\url{https://bop.felk.cvut.cz/datasets}} are 6D pose estimation benchmarks, consisting of 3D object models, training and test RGB/RGB-D images annotated with ground-truth 6D object poses and intrinsic camera parameters. In our work, we do not use the RGB-D data.
The LINEMOD dataset consists of 15 texture-less household objects with discriminative color, shape and size. Only 13 of the objects have the CAD models, so, following standard practice, we focus on them. Each object is associated with a training/testing image set showing one annotated object instance with significant clutter but only mild occlusion. Following~\cite{eric_uncertainty}, we split the data into a training set containing around 200 images per object and a test set containing around 1000 images per object. Occluded-LINEMOD provides additional ground-truth annotations for all modeled objects in one of the test sets from LINEMOD. This introduces challenging test cases with various levels of occlusion. Note that we use the real images from LINEMOD together with the synthetic ones provided with the dataset and generated using physically-based rendering~\cite{hlmo2020bop}. In our work, we respect the terms and conditions of use listed on the websites. 

\textbf{YCB-V}~\cite{xiang2017posecnn}\footnote{~\url{https://rse-lab.cs.washington.edu/projects/posecnn}} is a large-scale video dataset for 6D object pose estimation, which provides accurate 6D poses of 21 objects observed in 92 videos, with in total of 133,827 frames. The objects are highly occluded. 

\textbf{WDRNet}~\cite{hu2021wdrpose}\footnote{~\url{https://github.com/cvlab-epfl/wide-depth-range-pose}} and \textbf{ZebraPose}~\cite{su2022zebrapose}\footnote{~\url{https://github.com/suyz526/ZebraPose}} are open-source 6D pose estimation frameworks built in Pytorch~\cite{pytorch}, and are released under the non-commercial use license and MIT License, respectively. Together with WDRNet, we also exploit the detector pre-processing portion of the SO-Pose\cite{di2021SOpose} codebase\footnote{~\url{https://github.com/shangbuhuan13/SO-Pose}}, which is released under the Apache License 2.0. To implement and solve the Optimal Transport (OT) models in our method, we rely on the GeomLoss library~\cite{feydy2019interpolating}\footnote{~\url{https://github.com/jeanfeydy/geomloss}}, which falls under the MIT License. For the details of these licenses, please refer to the websites.

\textbf{Computing resources.} All experiments were conducted on an internal cluster, with Tesla V100 or A100 GPUs. All models were trained on one single GPU. 

\input{tables/supp_nkd}

\subsection{Hyper-parameters for Naive-KD and FKD}
In this section, as mentioned in the main paper, we provide the details of the hyper-parameter search for Naive-KD and FKD~\cite{zhangimproveFKD}. In both cases, this search was mostly focused on models with a DarkNet-tiny-H backbone and on 2 difficult LINEMOD classes, i.e., Ape and Duck.

\textbf{Naive-KD.} In the sparse 2D keypoints scenario, for WDRNet+, as shown in Table~\ref{tab:supp_nkd_tiny_h}, the best results are obtained with a norm $p=1$ and a distillation loss weight of 0.1, and with a norm $p=2$ with a weight of 0.1. We therefore provide the corresponding results for all classes and for the DarkNet-tiny-H and DarkNet-tiny backbones in Table~\ref{tab:supp_nkd}. Note that $p=2$ with a weight of 0.1 yields the best results for DarkNet-tiny-H, and $p=1$ with a weight of 0.1 gets the best performance for DarkNet-tiny. Therefore, we report the best results for each backbone in the main paper. Note that, for WDRNet+, Naive-KD hardly improves the student's performance. 

For ZebraPose, we use $p=1$ for the DarkNet-tiny student backbone, with a weight in \{0.1,1,10\}. As shown in Table~\ref{tab:supp_nkd_zebrapose}, a weight of 1.0 yields the best results. 

\input{tables/supp_fkd}

\textbf{FKD~\cite{zhangimproveFKD}.} We follow the same strategy as above, and report the results for Ape and Duck with FKD in Table~\ref{tab:supp_fkd_lm}. The best results are obtained with a distillation weight of 0.01. As the weight increases, the performance decreases significantly. We therefore adopted 0.01 as FKD weight for both the DarkNet-tiny-H and DarkNet-tiny backbones on the LINEMOD dataset. For FKD, we also conducted a hyper-parameter search on Occluded-LINEMOD. As shown in Table~\ref{tab:supp_fkd_occ}, a distillation weight of 0.01 also achieves the best results. Note that we did not test a weight of 0.1 on all classes because of the worse results it gave on Ape and Duck.

\input{tables/supp_ourkd}

\subsection{Hyper-parameters for our Approach}
In this section, we include the hyper-parameter search for our proposed keypoint distribution alignment distillation method, including the norm $p$ and the weight of our distillation loss. As for WDRNet+, we focused this search on DarkNet-tiny-H for the Ape and Duck classes. As shown in Table~\ref{tab:supp_ourkd_lm}, $p=2$ yields much better results than $p=1$, and we therefore use $p=2$ in the main paper. As for the loss weight, on the LINEMOD dataset, 5 yields the best results, which we use to report the results on the 13 classes in the main paper. For Occluded-LINEMOD, as shown in Table~\ref{tab:supp_ourkd_occ}, we obtain the best results with a weight of 0.1. Note that our preliminary experiments with a weight of 1 showed worse performance, and we thus did not compute full results with weights larger than 0.1.

For ZebraPose, as shown in Table~\ref{tab:supp_ourkd_occ_zebrapose}, we observed a weight of 1.0 to only yield a marginal improvement on the class Ape. We therefore increased the weight to 10.0 and 100.0, both of which led to higher improvements on Ape. These improvements also materialized on the other classes. Thus, in the main paper, we report the results with a weight of 100.0.

%% file: tables/supp_nkd.tex
\begin{table}[!t]
    \centering
    \captionof{table}{\textbf{Results of Naive-KD with DarkNet-tiny-H backbone on Ape and Duck with WDRNet+.} We report the ADD-0.1d for the Naive-KD with $p=1$ and $p=2$.}
    \label{tab:supp_nkd_tiny_h}
    \vspace{0.1cm}
    \resizebox{0.99\linewidth}{!}{
    \begin{tabular}{r |  c  |  c | c c c | c c c }
    \toprule
     \multirow{2}{*}{Class} &  \multirow{2}{*}{Teacher} & \multirow{2}{*}{Student} & \multicolumn{3}{c|}{$p=1$}  &\multicolumn{3}{c}{$p=2$}  \\
     \cmidrule{4-6} \cmidrule{7-9}
     &   &  & 0.01 & \textbf{0.1 }& 1.0  & 0.01 & \textbf{0.1} & 1.0  \\
    \midrule
    Ape         & 82.6 & 65.4 & 63.2    & 64.4 & 65.7  & 63.8 & 64.1 & 64.8 \\
    Duck        & 76.0 & 64.3 & 59.4    & 63.3 & 60.3 & 59.0 & 63.6 & 62.2 \\
    \midrule
    AVG.        & 79.3 & 64.8 & 61.3    & \textbf{63.9} & 63.0   & 61.4 & \textbf{63.9} & 63.5  \\
    \bottomrule
    \end{tabular}
    }

    \centering
    \vspace{5mm}
       \captionof{table}{\textbf{Results of Naive-KD on LINEMOD dataset with WDRNet+.} We report the ADD-0.1d for the Naive-KD with DarkNet-tiny-H and DarkNet-tiny backbones with the different norms $p$ and the weights searched from Table~\ref{tab:supp_nkd_tiny_h}.}
     \label{tab:supp_nkd}
    \vspace{0.1cm}
    \resizebox{0.99\linewidth}{!}{
    \begin{tabular}{r |  c  |  c c c | c c c  }
    \toprule
    \multirow{3}{*}{Class} &  \multirow{3}{*}{Teacher} &   \multicolumn{3}{c|}{DarkNet-tiny-H}  &\multicolumn{3}{c}{DarkNet-tiny}  \\
     \cmidrule{3-8}
        &   & \multirow{2}{*}{Student} & $p=1$  &\textbf{$p=2$}& \multirow{2}{*}{Student} & \textbf{$p=1$}  &$p=2$   \\
        &   &                          & 0.1   &  \textbf{0.1} & & \textbf{0.1 }  &  0.1 \\
    \midrule
    Ape         & 82.6 & 65.4 & 64.4 & 64.1 & 73.4 & 74.1 & 74.0 \\
    Bvise       & 95.5 & 92.0 & 90.6 & 91.4 & 95.2 & 95.4 & 96.6 \\
    Cam         & 93.8 & 78.4 & 77.8 & 79.1 & 91.2 & 89.7 & 90.0 \\
    Can         & 95.7 & 82.2 & 78.7 & 81.0 & 94.4 & 92.7 & 92.9 \\
    Cat         & 92.0 & 81.5 & 77.8 & 78.7 & 87.2 & 85.0 & 82.0 \\
    Driller     & 94.8 & 85.5 & 87.6 & 87.4 & 92.2 & 93.1 & 93.2 \\
    Duck        & 76.0 & 64.3 & 63.3 & 63.6 & 70.9 & 74.4 & 73.9 \\
    Eggbox$^*$  & 99.1 & 95.8 & 95.3 & 95.0 & 99.3 & 98.7 & 99.4 \\
    Glue$^*$    & 96.4 & 90.7 & 92.6 & 91.2 & 97.2 & 97.1 & 96.9 \\
    Holep       & 86.2 & 73.2 & 71.6 & 72.3 & 78.0 & 82.1 & 81.0 \\
    Iron        & 93.6 & 86.3 & 86.4 & 86.3 & 92.1 & 92.1 & 91.9 \\
    Lamp        & 97.7 & 93.6 & 93.3 & 94.2 & 96.6 & 95.3 & 96.5 \\
    Phone       & 91.2 & 76.0 & 75.7 & 75.8 & 87.5 & 88.4 & 87.4 \\
    \midrule
    AVG.        & 91.9 & 81.9 & 81.2 & \textbf{81.6} & 88.9 & \textbf{89.1} & 88.9 \\
    \bottomrule
    \end{tabular}
    }
    \centering
    \vspace{5mm}
    \captionof{table}{\textbf{Weight searching for Naive-KD on OCC-LINEMOD dataset with ZebraPose (Ape and Duck).} We report the ADD-0.1d for Naive-KD with different weights.}
    \label{tab:supp_nkd_zebrapose}
    \vspace{0.1cm}
    \resizebox{0.8\linewidth}{!}{
    \begin{tabular}{r |  c  |  c | c c c c  }
    \toprule
     Class &  Teacher &  Student & 0.1  & \textbf{1.0} & 10.0\\
    \midrule
    Ape     & 57.9   & 47.2  &	49.1 &	51.1  & 50.5\\
  
    Duck    & 64.5  & 57.2  &	57.4 &	60.7  & 59.7 \\

    \midrule
    AVG.    & 61.2	& 52.2	& 53.3	 & \textbf{55.9} & 55.1 \\

    \bottomrule
    \end{tabular}
    }
\end{table}

%% file: tables/supp_fkd.tex
\begin{table}[!t]
    \centering
    \captionof{table}{\textbf{Weight searching for FKD on LINEMOD dataset with WDRNet+ (Ape and Duck).} We report the ADD-0.1d for FKD~\cite{zhangimproveFKD} with different weights.}
    \label{tab:supp_fkd_lm}
    \vspace{0.1cm}
    \resizebox{0.9\linewidth}{!}{
    \begin{tabular}{r |  c  |  c | c c c c  }
    \toprule
     Class &  Teacher &  Student & 0.001 & \textbf{0.01} & 0.1  & 1.0 \\
    \midrule
    Ape         & 82.6 & 65.4 & 66.5 & 68.4 & 66.5 & 65.0 \\
  
    Duck        & 76.0 & 64.3 & 65.2 & 66.8 & 61.2 & 60.3 \\

    \midrule
    AVG.        & 79.3 & 64.8 & 65.9 & \textbf{67.6} & 63.8 & 62.7 \\

    \bottomrule
    \end{tabular}
    }

    \centering
    \vspace{5mm}
       \captionof{table}{\textbf{Results of FKD on OCC-LINEMOD dataset with WDRNet+.} We report the ADD-0.1d for FKD~\cite{zhangimproveFKD} with different weights. Note that due to the worse results on Ape and Duck with a weight of 0.1, we didn't extend this setting to other classes.}
    \label{tab:supp_fkd_occ}
    \vspace{0.1cm}
    \resizebox{0.9\linewidth}{!}{
    \begin{tabular}{r |  c  |  c |  c c c }
    \toprule
     Class      & Teacher  & Student  &  0.001   &  \textbf{0.01}  & 0.1 \\
    \midrule
    Ape         & 33.4 & 25.5 & 26.8 & 26.7 & 22.6\\
    Can         & 70.9 & 46.6 & 52.8 & 53.9 & - \\
    Cat         & 45.1 & 31.4 & 31.0 & 31.1 & - \\
    Driller     & 70.9 & 51.2 & 52.3 & 52.1 & - \\
    Duck        & 27.0 & 22.5 & 24.7 & 25.3 & 19.8\\
    Eggbox$^*$  & 53.7 & 43.4 & 47.9 & 49.0 & - \\
    Glue$^*$    & 70.7 & 54.5 & 54.3 & 55.6 & - \\
    Holep       & 59.7 & 49.3 & 51.0 & 52.2 & - \\
    \midrule
    AVG.        & 53.9 & 40.5 & 42.6 & \textbf{43.2} & - \\
    \bottomrule
    \end{tabular}
    }
\end{table}

%% file: tables/supp_ourkd.tex
\begin{table}[t]
    \centering
    \captionof{table}{\textbf{Results of our proposed KD with DarkNet-tiny-H backbone on LINEMOD dataset (Ape and Duck) with WDRNet+.} We report the ADD-0.1d for our proposed KD with different  $p$s and weights.}
    \label{tab:supp_ourkd_lm}
    \vspace{0.1cm}
    \resizebox{0.98\linewidth}{!}{
    \begin{tabular}{r |  c  |  c | c c  | c c c }
    \toprule
    \multirow{2}{*}{Class} &  \multirow{2}{*}{Teacher} & \multirow{2}{*}{Student} & \multicolumn{2}{c|}{$p=1$}  &\multicolumn{3}{c}{$p=2$}  \\
     \cmidrule{4-8}
            &           &   & 1.0 & 10.0 & 1.0 & \textbf{5.0} & 10.0 \\
    \midrule
    Ape         & 82.6 & 65.4 & 61.9    & 61.5  & 66.5  & 69.4 & 67.0 \\
    Duck        & 76.0 & 64.3 & 61.2    & 61.9  & 65.1  & 66.5 & 65.8 \\
    \midrule
    AVG.        & 79.3 & 64.8 & 61.6 & 61.7 & 65.8 & \textbf{67.9} & 66.4 \\
    \bottomrule
    \end{tabular}
    }

    \centering
    \vspace{5mm}
     \captionof{table}{\textbf{Results of our proposed KD on OCC-LINEMOD dataset with WDRNet+.} We report the ADD-0.1d for our proposed KD with different weights.}
    \label{tab:supp_ourkd_occ}
    \vspace{0.1cm}
    \resizebox{0.78\linewidth}{!}{
    \begin{tabular}{r |  c  |  c |  c c  }
    \toprule
     Class      & Teacher  & Student  &  0.01   &  \textbf{0.1}  \\
    \midrule
    Ape         & 33.4 & 25.5 & 23.5 & 25.7 \\
    Can         & 70.9 & 46.6 & 51.2 & 53.5 \\
    Cat         & 45.1 & 31.4 & 31.3 & 32.2 \\
    Driller     & 70.9 & 51.2 & 51.5 & 52.9 \\
    Duck        & 27.0 & 22.5 & 20.0 & 25.7 \\
    Eggbox$^*$  & 53.7 & 43.4 & 47.9 & 48.2 \\
    Glue$^*$    & 70.7 & 54.5 & 54.3 & 55.8 \\
    Holep       & 59.7 & 49.3 & 51.0 & 52.1 \\
     \midrule
    AVG.        & 53.9 & 40.5 & 41.3 & \textbf{43.2} \\
    \bottomrule
    \end{tabular}
    }
    \centering
    \vspace{5mm}
     \captionof{table}{\textbf{Results of our proposed KD on OCC-LINEMOD dataset with ZebraPose.} We report the ADD-0.1d for our proposed KD with different weights.}
    \label{tab:supp_ourkd_occ_zebrapose}
    \vspace{0.1cm}
    \resizebox{0.9\linewidth}{!}{
    \begin{tabular}{r |  c  |  c |  c c  c}
    \toprule
    Class   & Teacher   & Student   & 1.0        & 10.0    & \textbf{100.0}  \\
     \midrule
    Ape     & 57.9      & 47.2      & 47.9      & 50.1     & \textbf{52.0} \\
    Can     & 95.0      & 93.2      & -         & 94.0     & \textbf{94.2} \\
    Cat     & 60.6      & 53.1      & -         & 54.8     & \textbf{55.2} \\
    Driller & 94.8      & 90.3      & -         & 89.1     & \textbf{90.4} \\
    Duck    & 64.5      & 57.2      & -         & 60.8     & \textbf{61.0} \\
    Eggbox* & 70.9      & 69.6      & -         & 70.4     & \textbf{70.7} \\
    Glue*   & 88.7      & 84.1      & -         & 84.3     & \textbf{84.3} \\
    Holep   & 83.0      & 75.8      & -         & 76.9     & \textbf{78.8} \\
    \midrule
    AVG.    & 76.9     & 71.4       & -         & 72.5      & \textbf{73.3} \\
    \bottomrule
    \end{tabular}
    }
    
\end{table}